\documentclass[10pt,twocolumn,letterpaper]{article}

\usepackage{iccv}
\usepackage{times}
\usepackage{epsfig}
\usepackage{graphicx}
\usepackage{amsmath}
\usepackage{amssymb}
\usepackage{dirtytalk}

\usepackage{subcaption}
\usepackage{array}
\usepackage{algpseudocode}
\usepackage{soul}
\usepackage[pagebackref=true,breaklinks=true,letterpaper=true,colorlinks,bookmarks=false]{hyperref}
\usepackage[font=footnotesize,labelfont=bf]{caption}

\iccvfinalcopy 

\def\iccvPaperID{14} 

\ificcvfinal\fi
\begin{document}

\title{An empirical study of the relation between network architecture and  complexity}

\author{
\begin{tabular}[t]{c@{\extracolsep{6em}}c}
    \hspace{5mm} Emir Konuk & Kevin Smith \\
    \multicolumn{2}{c}{KTH and Science for Life Laboratory, Stockholm} \\
    \multicolumn{2}{c}{\tt\small \{ekonuk, ksmith\}@kth.se}
\end{tabular}}

\maketitle

\begin{abstract}
In this preregistration submission, we propose an empirical study of how networks handle changes in complexity of the data. We investigate the effect of network capacity on generalization performance in the face of increasing data complexity. For this, we measure the generalization error for an image classification task where the number of classes steadily increases. We compare a number of modern architectures at different scales in this setting. The methodology, setup, and hypotheses described in this proposal were evaluated by peer review before experiments were conducted.

\end{abstract}

\section{Introduction}
\label{sec:introduction}

The complexity of a learning task is one of the most important determinants of how well a model performs, yet relatively little is understood about the effects of data complexity in a practical setting. Although we lack a rigorous definition of complexity, many agree that certain factors contribute to the complexity in a classification problem including: the dataset size in relation to the dimensionality of the data, the intrinsic ambiguity of the classes, and how compactly the decision boundary can be expressed~\cite{ho2002complexity}.

The capacity of a network describes the complexity of the functions it can potentially model. Naturally, data with high complexity require a model with high effective capacity. In recent findings counter to the conventional wisdom, Zhang et al.~\cite{rethinking} showed that high effective capacity models can both memorize and generalize well whereas Neyshabur et al.~\cite{exploring_generalization} showed that networks with higher capacity also generalize better.\footnote{We are concerned with a low generalization error, not a small generalization gap (the difference between test and training error).}

\begin{figure}[t]
    \centering
    \hspace*{-0.5cm} 
    \begin{tabular}{cc}
    \includegraphics[width=0.44\linewidth]{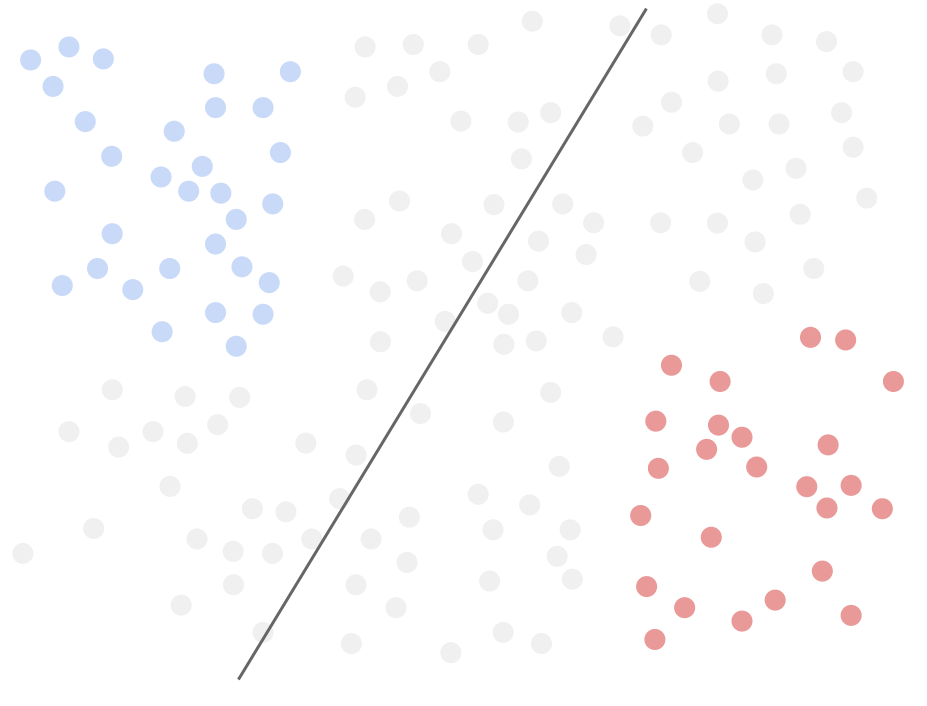}&
    \includegraphics[width=0.44\linewidth]{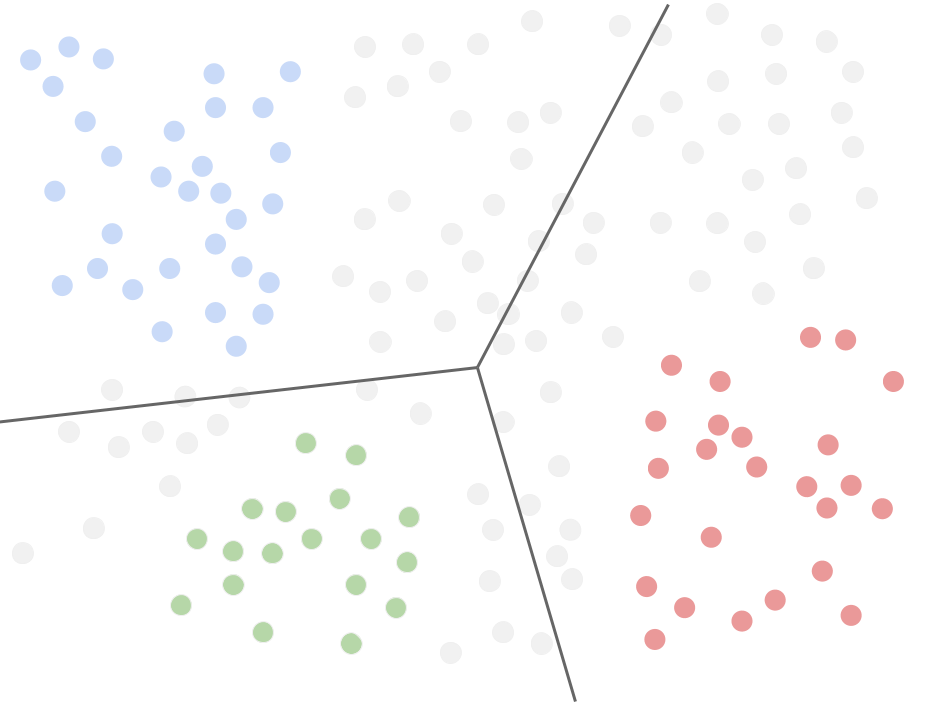}\\
    
    \footnotesize{$S_{2}$ (2 classes)} &
    \footnotesize{$S_{3}$ (3 classes)} \\
    \footnotesize{lowest complexity} & \\
    
     \includegraphics[width=0.44\linewidth]{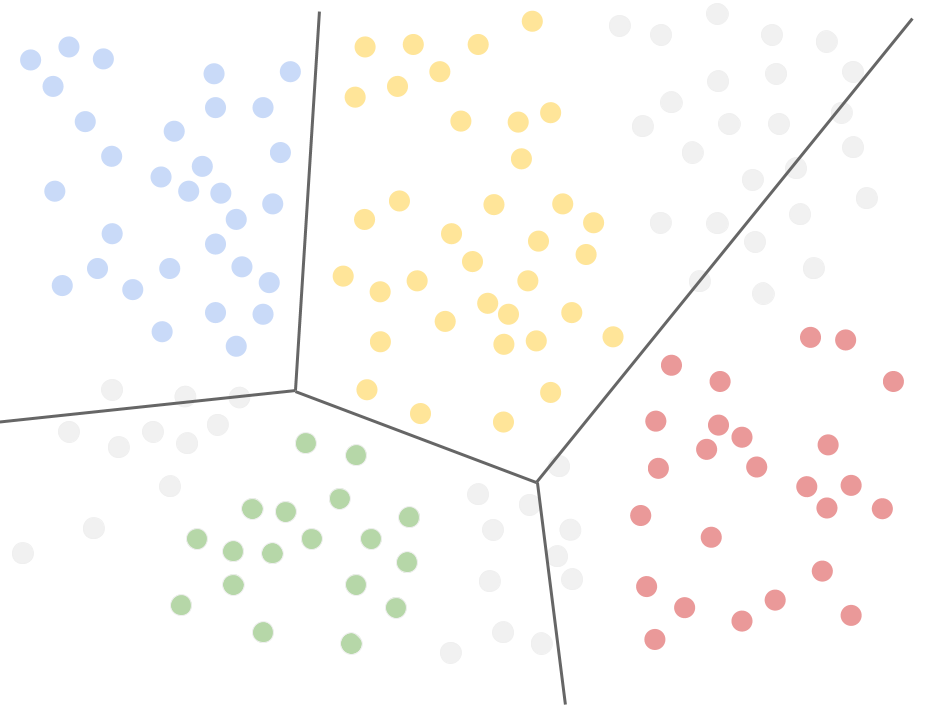}&
       \includegraphics[width=0.44\linewidth]{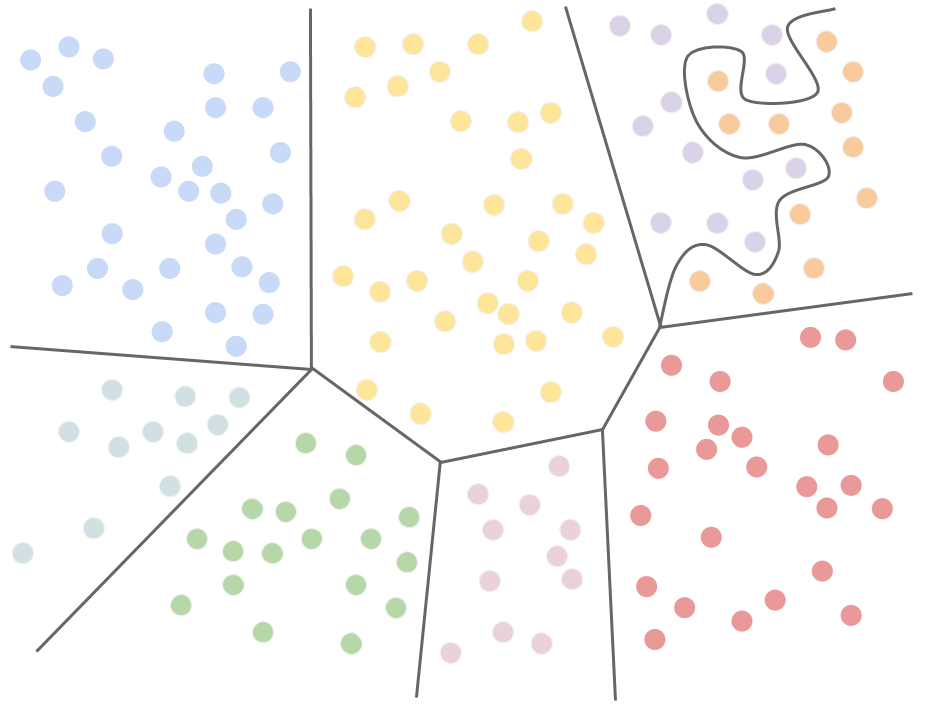}\\
       \footnotesize{$S_{4}$  (4 classes)} &
       \footnotesize{$S_{K}$  ($K$ classes)}\\
      & \footnotesize{highest complexity}\\
    \end{tabular}
    \footnotesize{\caption{One way to increase complexity is to add classes to the classification task. A training dataset with few classes has a simpler decision function than a dataset with many classes. We form a low-complexity dataset by sampling from a dataset containing many classes to create a subset with only 2 classes, $S_{k=2}$. Another class is sampled and added to the subset $S_{3}$, and this process is repeated for $k \leq K$ classes. We attempt to avoid sampling classes with strong ambiguity (e.g. the purple and orange classes) by inspecting the confusion matrix of a fully trained deep model. Classes that are easily confused with those already in $S_{k}$ are not sampled.}
    \label{fig:complexity}}
    \vspace{-4mm}
\end{figure}

In this paper,~\textit{we perform an empirical study to characterize how generalization error relates to network capacity when the complexity of the data changes}.
Our aim is to improve our understanding of the capacity of various deep neural architectures and potentially help guide the design process. Instead of utilizing theoretical bounds to calculate an architecture's capacity~\cite{harvey2017nearly} or measures based on the norms of network parameters for explaining a trained network's generalization performance~\cite{exploring_generalization}, we take an empirical approach and treat the generalization error on a common image classification problem as an indicator for effective network capacity. Starting from a very simple, low complexity problem, we repeatedly calculate the generalization error of architectures with different effective capacities on increasingly complex data (Fig.~\ref{fig:complexity}).
While the exact complexity of the data cannot be controlled, we ensure monotonic increases in complexity by repeatedly introducing new classes to the classification task. 

Our empirical analysis will address the following questions about convolutional neural networks (CNNs)
:
\begin{itemize}
    \item How can we characterize the changes in generalization error as complexity is increased?
    \item How does capacity relate to generalization error as complexity increases? Does scaling the network by a particular dimension (e.g. depth) offer an advantage?
    \item How do architectural innovations such as group convolutions affect the capacity/generalization/complexity relationship?
\end{itemize}

In a realistic training setting with measures to avoid overfitting (\textit{i.e.}~regularization, batch norm, early stopping) we hypothesize that the generalization error will follow a similar trajectory to the black curve shown in Fig.~\ref{fig:hypothesis} as the complexity increases.
When the number of classes is small and the data complexity is low, the capacity of the network allows for generalization to approach the irreducible error. As the data complexity increases, we expect a faster increase in the error as the relative capacity of the network becomes insufficient to generalize well.
In the final regime of very complex datasets, as in \textit{extreme classification} with over 10,000 classes,
generalization error will saturate.

\section{Related work}
\label{sec:related}
Hestness et al.~\cite{hestness2017deep} empirically showed that the generalization error of deep networks scale with a power-law with respect to the amount of data. These results are useful in estimating how much data and computation is required for a task. However, their study does not address how generalization is affected by the complexity of the data.\\

\noindent\textbf{Complexity measures:} Direct measures of the complexity of a problem such as the Kolmogorov measure are infeasible to compute in most cases, and though approximate metrics have been suggested, they are not useful for deep networks \cite{lorena2018complex}. Various theoretical studies prove the \textit{bounds} on the capacity of neural networks ranging from VC complexity 
to 
bounds for fully connected ReLU networks \cite{harvey2017nearly}. 

\begin{figure}[t]
    \centering
    \hspace*{-0.5cm} 
    \includegraphics[width=1.1\linewidth]{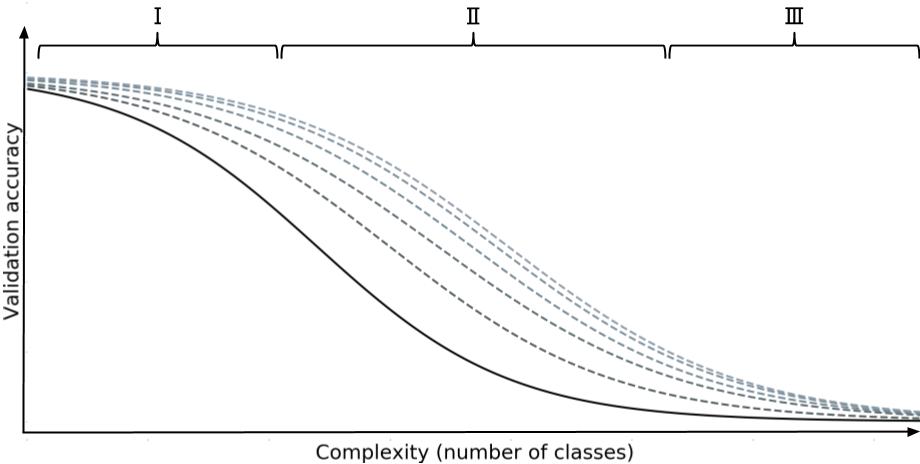}
    \caption{Validation accuracy vs. data complexity. Regions I, II and III are under-complex, complex, and over-complex relative to the model's capacity. Validation accuracy approaches saturates when the data is under-complex, starts to break down in the complex region, and deteriorates to a random classifier when data is over-complex. The black curve represents expected performance of the base model, and the dashed curves represent scaled models with higher capacity (approximately constant scaling factor). We expect the improvement due to scaling to have diminishing returns, and expect to find less benefit from scaling in regions I and III.}
    \label{fig:hypothesis}
    \vspace{-4mm}
\end{figure}

Even though these methods provide insights into the relationship between complexity and architecture, they do not help much in choosing an architecture or how to scale it to have lower generalization error. In this direction, Gus et al.~\cite{guss2018characterizing} used persistent homology to characterize different architectures' generalization capabilities. They created increasingly more complex datasets, as measured by persistent homology, and compared the performances of different architectures on these datasets. Unfortunately, this approach also has limited practical use as persistent homology is not feasible to compute for high dimensional data. 

In this paper, we forego attempts to directly measure dataset complexity. Instead, we \textit{rank} datasets by complexity using a relatively simple heuristic.\\

\noindent\textbf{Network scaling:} Tan et al.~\cite{tan2019efficientnet} report that when scaling a network, accuracy suffers from diminishing returns. They propose a compound scaling scheme to mitigate the effect to some degree. Note that their analysis  focuses on a dataset with static complexity, corresponding to a vertical line in region II of Fig~\ref{fig:hypothesis}. We investigate how the effect of diminishing returns changes when data complexity changes, covering the $x$-axis in Fig~\ref{fig:hypothesis}.

\section{Methodology and experimental protocol}
\label{sec:Protocol}

In order to assess the effect of capacity, we measure the generalization error of a modest sized network starting from a very simple, low complexity problem. As we gradually increase the problem complexity, we expect the generalization error to increase as well, reflecting the reduced capacity of the network to generalize. By repeating the experiment with larger, scaled networks and different architectures, we empirically investigate the effect of architecture on the generalization capacity.   
The main experiment is to train a network on a subset of ImageNet, where the number of classes gives an indication of complexity. The networks tested will be scaled versions of small base networks. The experiment will be repeated multiple times:
\begin{itemize}
    \item on increasingly more complex data subsets,
    \item with different scaling factors along different dimensions (width, depth, cardinality, growth parameter),
    \item with a different base network architecture (vanilla CNN, ResNet, ResNeXt, DenseNet).
\end{itemize}

\noindent \textbf{Base networks:} We test five different base models. The first two are based on ResNet \cite{resnet}, which defines two types of convolutional blocks:~basic and bottleneck blocks. For our experiments, these two block types are treated as different base network architectures. We also use ResNeXt \cite{resnext} as a base architecture block. All residual base models have 10 convolutional layers. DenseNet \cite{densenet} is included as another base model, with 85 layers. Finally, a vanilla CNN without any skip connections is included.  Base model comparisons are only made between these five, for networks with similar number of total parameters.\\

\noindent\textbf{Generating increasingly complex data sets:}
To examine the network's response to complexity, we generate datasets with different numbers of classes, as depicted in Fig. \ref{fig:complexity}. We sample subsets of ImageNet to form these working datasets. Starting from a low-complexity subset $S_{k=2}$ with two randomly sampled classes, we repeatedly sample a class $i$ from ImageNet's $>$20,000 classes without replacement, and add the corresponding set of examples $E_{i}$ to the working set $S_{k+1} = S_{k} \cup E_{i}$. In this manner, we create subsets of ImageNet \{$S_{2}, \ldots, S_{k}, \ldots, S_{K}\}$  with exponential increases in number of classes, and by proxy, complexity. 

The addition of each class introduces some unknown amount of complexity to the data. This presents two problems. First, if the subsets are not consistent between experiments, comparisons are difficult. To address this we keep the series of subsets constant for each experiment (classes added in the same order). The second problem is that adding randomly sampled classes introduces an inconsistent amount of complexity at each step based on semantic density~\cite{deng2010does}. While we cannot strictly control the \textit{amount} of complexity added at each step, we would like to avoid adding overly complex steps (e.g. ambiguous classes). To accomplish this, we use importance sampling to add \textit{easier} classes first, and stop at $K$ classes once a certain ambiguity is reached, determined by the aggregate confusion matrices of ResNet152, DenseNet201 and ResNeXt101 models trained on full dataset. Confusing classes are sampled with less probability when growing the working set. We repeat the training set generation procedure multiple times and repeat all experiments to minimize effects of class ordering.

To control for confounding effects from increased data size in the previous setup, we conduct a second set of experiments using an alternative method for increasing classes, in which the total number of training samples is fixed ($N\simeq50,000$ samples). Starting from $\sim$50 classes, we introduce new classes by replacing some of the existing samples while keeping the training set balanced.\\

\noindent\textbf{Network scaling:}
We consider two main kinds of network scaling: width and depth. To increase depth, we add more convolutional layers to the network before the downsampling layers. We keep the number of channels and feature size constant, \textit{i.e.}, we keep the number of blocks in a network constant when adding new layers. The scaling schedule will be determined through an initial parameter search described below. The standard method to increase a network's width is to simply create more feature channels. The width scaling schedule is determined through an initial parameter search. For ResNeXt, the scaling method is to increase cardinality, i.e. the number of grouped convolutions in each ResNeXt block. For DenseNet, the growth parameter and number of layers are scaled in tandem.\\

\noindent\textbf{Training:} For training, we use an Adam optimizer along with cyclical learning rates \cite{smith2015cyclical} to simplify hyperparameter selection. Early stopping and various data augmentations, e.g. random crop, flip, rotation, color, brightness, noise are employed for regularization.\\ 

\noindent\textbf{Initial parameter search:} We do a hyperparameter search using the base model to choose the number of classes to add per subset at each growth step. For determining scaling schedules, we first find a suitably large subset so that the error of a modest sized network is in region II of Fig. \ref{fig:hypothesis}. Then we do a hyperparameter search to determine the scaling schedule that ends up near region III.

\footnotesize{\noindent\textbf{\\Acknowlegements:} This work was supported by WASP-AI and the Swedish Research Council (VR) 2017-04609.
We also thank C. Matsoukas for helpful discussions.
}

\begin{figure*}[ht]
\begin{tabular}{cc}
\includegraphics[width=0.5\linewidth]{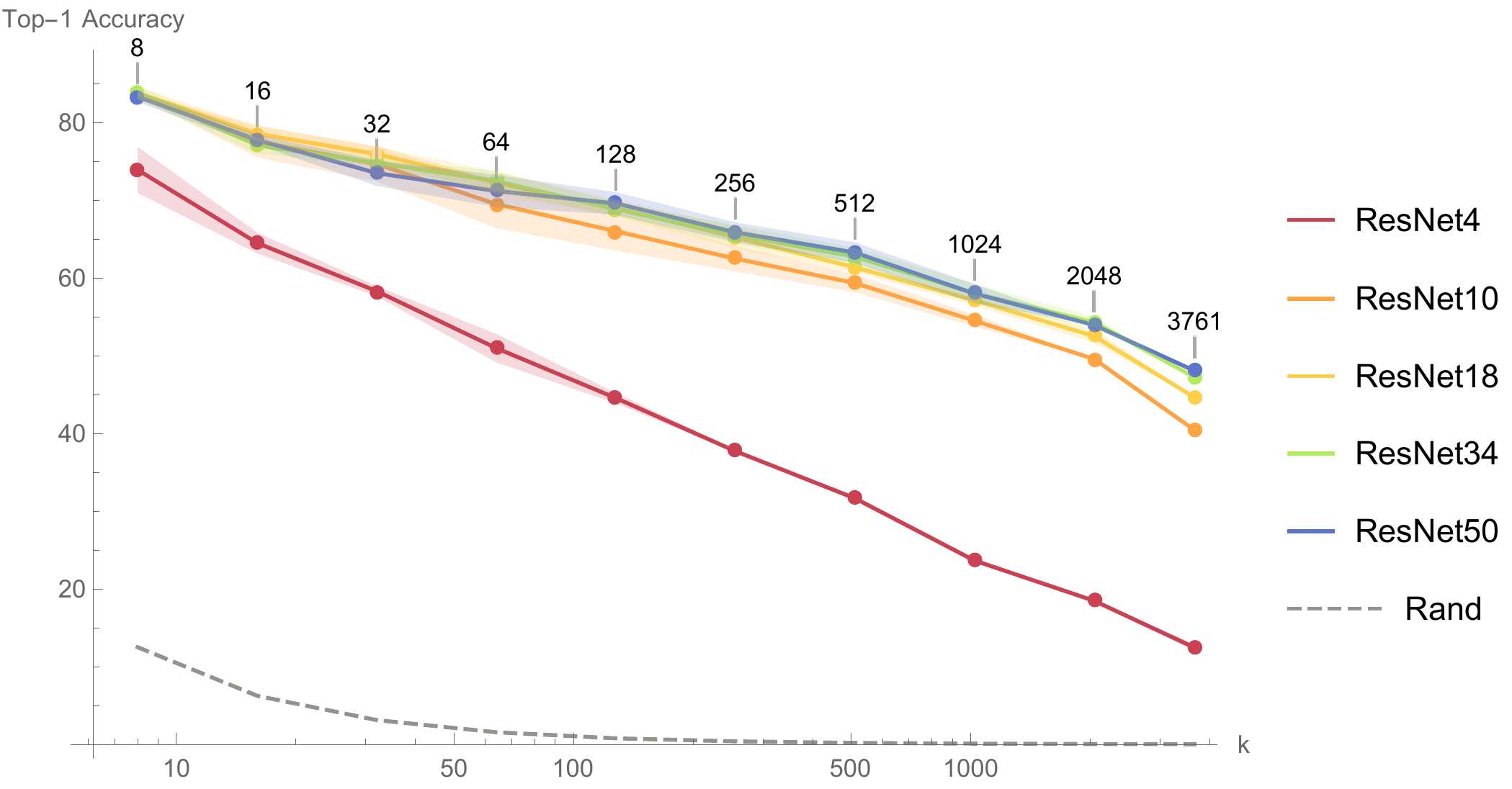} &
\includegraphics[width=0.5\linewidth]{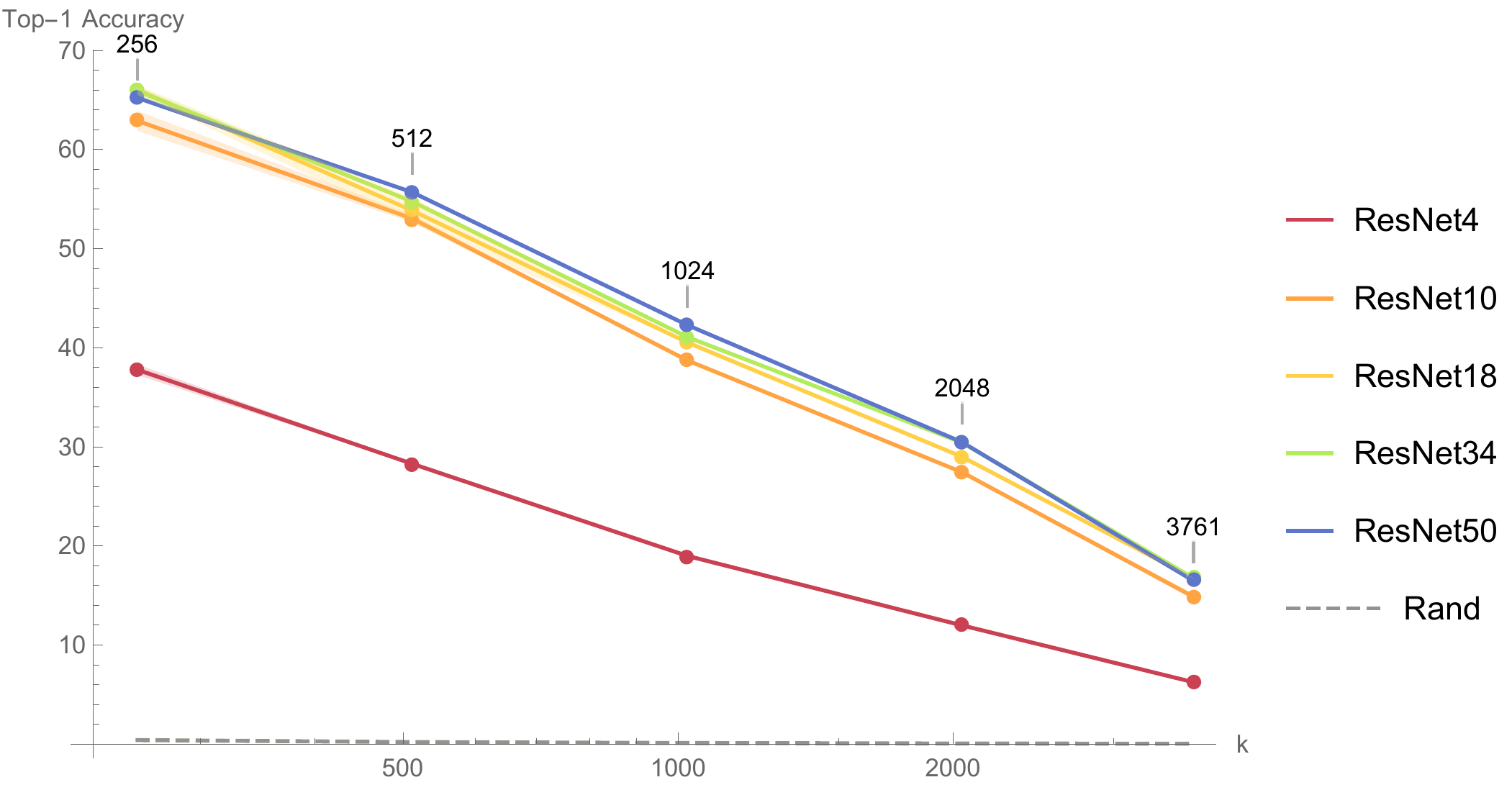} \\
\footnotesize{(a) Depth vs. complexity, $N$ increases with $k$} &
\footnotesize{(b) Depth vs. complexity, $N$ fixed to 100,000}\\
\includegraphics[width=0.5\linewidth]{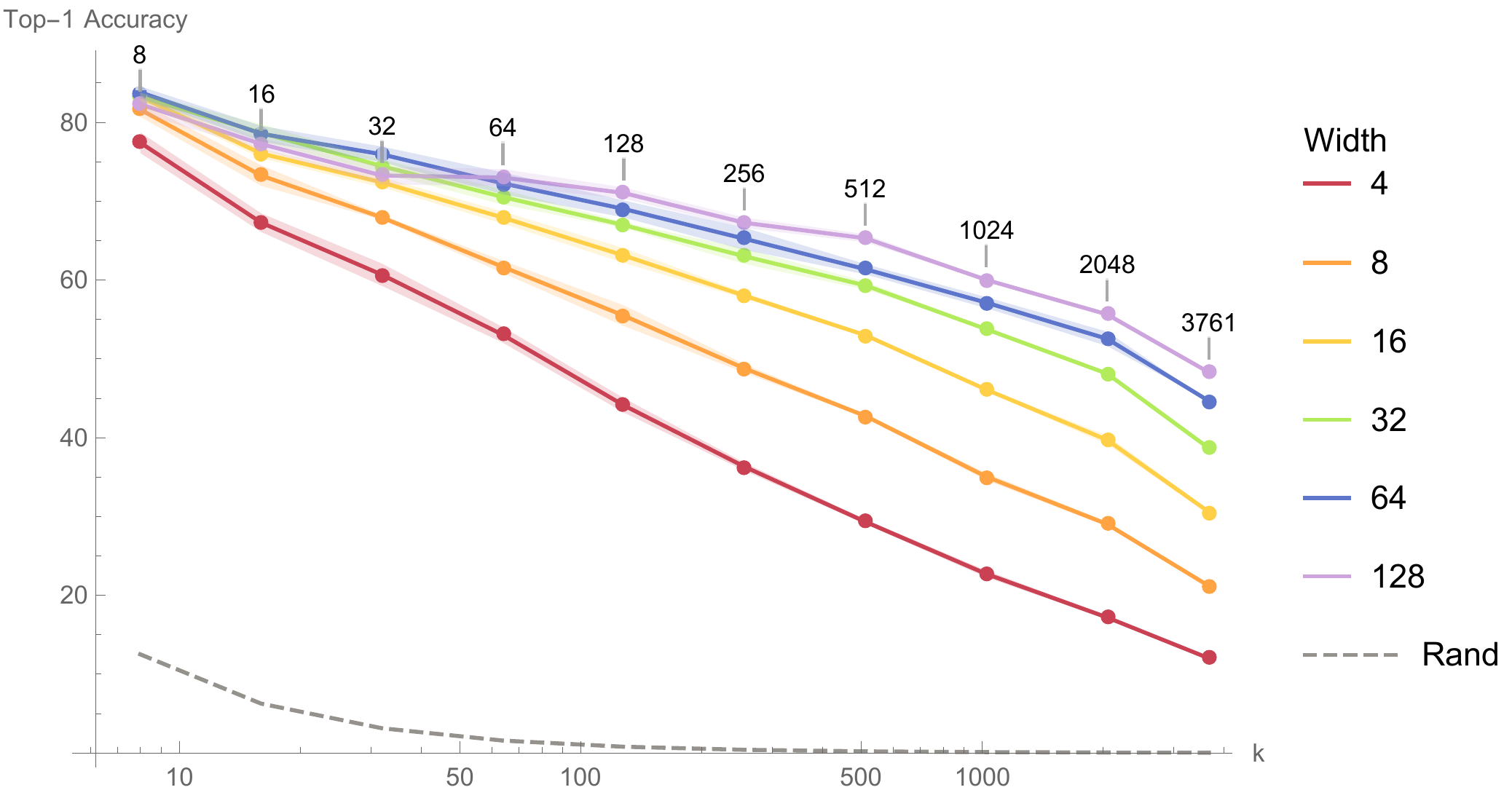} &
\includegraphics[width=0.5\linewidth]{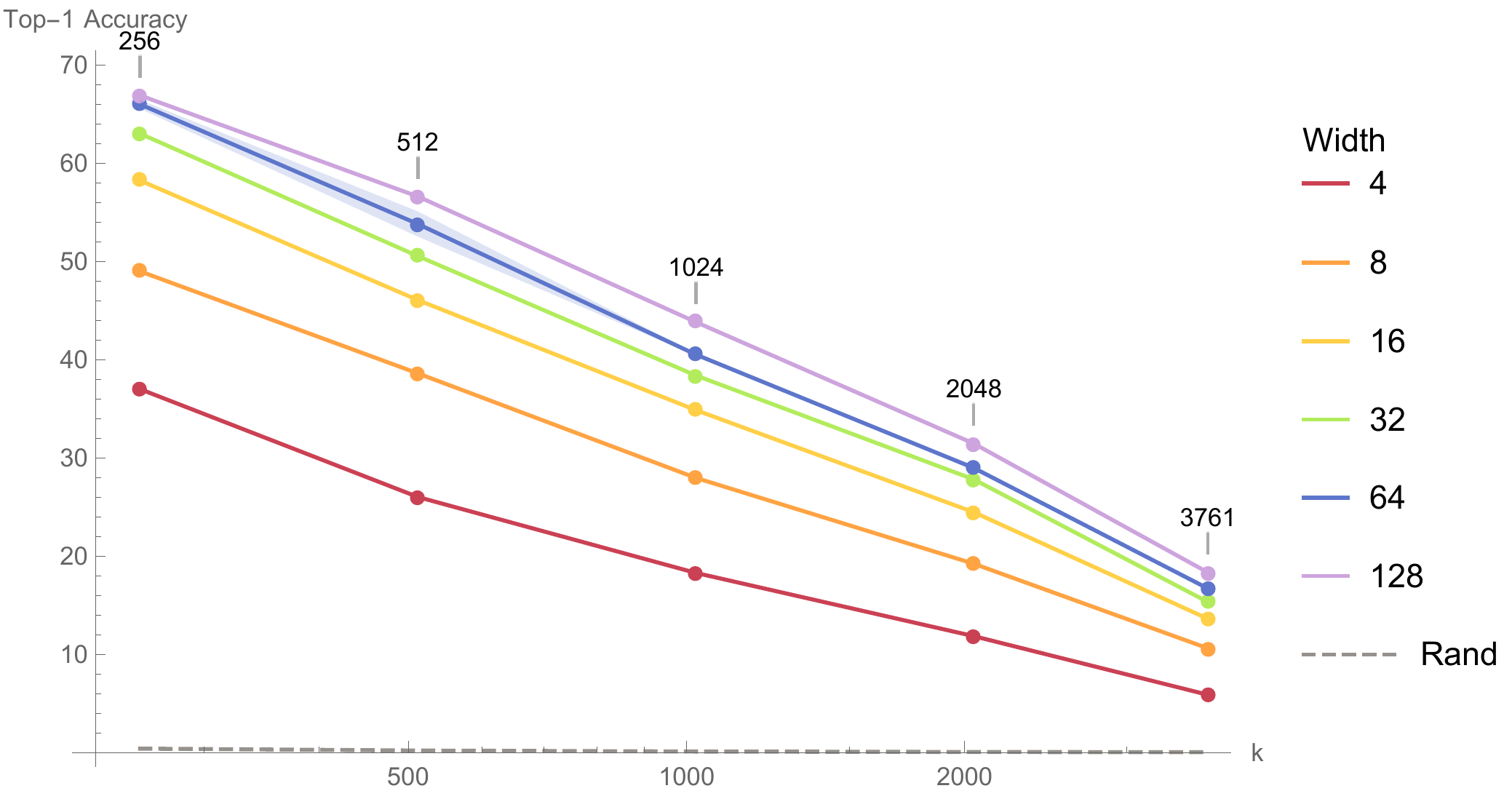} \\
\footnotesize{(c) Width vs. complexity, $N$ increases with $k$} &
\footnotesize{(d) Width vs. complexity, $N$ fixed to 100,000}
\end{tabular}
\caption{Effects of increasing data complexity on architectures with different capacity. Top-1 image classification validation accuracy for networks with various depths (a,b) and widths (c,d) as the number of classes $k$ in the task increases ($x$-axis is scaled in $\log_{10}$). In (a,c), the number of training examples $N$ grows with the number of classes, in (b,d) $N$=100,000. Image data was sampled from the broader ImageNet repository, not the ILSVRC subset, and includes up to 3,761 classes with $\approx$400 training samples per class. Results shown are the mean top-1 accuracy for basic ResNet blocks \cite{resnet} with up to 5 repetitions (some repetitions were not completed). Shaded areas indicate 1 s.d. The dashed lines indicate performance of a random classifier for reference.} 
\label{fig:Results}
\end{figure*}
{\small
\bibliographystyle{unsrt}
\bibliography{egbib}
}

\normalsize

\section{Results}
\label{sec:Results}
Our empirical findings roughly align with our hypothesis as it was outlined in Fig.~\ref{fig:hypothesis}. Validation accuracy decreases with increasing problem complexity. Meanwhile, increasing the model size improves accuracy for all levels of complexity. Addressing our main questions, empirical results indicate that \textit{the relationship between validation accuracy and complexity can be characterized as log-linear}, as shown in Fig.~\ref{fig:Results}. The slope of this log-linear curve is determined by the size of the models. Smaller models have a steeper performance drop. It is worth repeating that complexity is guaranteed to increase along the $x$-axis, but not necessarily linearly. This may partly explain the dips in the rightmost part of some curves. 

We were unable to complete experiments investigating our third question, how architectural innovations affect the generalization/complexity relationship. We leave this for future work.

\noindent\textbf{Final experimental setup:} We note the following deviations from our planned experiments. Due to time and resource constraints, we limited our investigation to basic ResNet architectures using a single class ordering in the training data, and some of the five planned repetitions in Fig.~\ref{fig:Results} were not completed. The ResNet4 contains only one basic residual block. To push the bounds of model capacity, we generate a dataset with 3,761 classes sampled from the 
broader ImageNet repository. We used SGD with momentum with warm-up. Models were trained with mixed precision for 90 epochs, gradually decreasing the learning rate \textit{without early stopping}. For each repetition, the initial learning rates were picked randomly from the learning rate interval just before the divergence point in the learning rate range test \cite{smith2015cyclical}. The batch size was kept constant at 128 for all experiments. 

\vspace{-1mm}
\section{Conclusion}
\label{sec:Conclusion}
In this work, for a limited but practical setting, we show that a deep network's performance decreases log-linearly with problem complexity as indicated by the number of categories. Our results are in support of earlier findings by Zhang et al. \cite{rethinking}, that even though a deep network may memorize the dataset, it still generalizes well. We also expand the findings of Hestness et al. \cite{hestness2017deep} by showing that scaling the model size has diminishing returns for problems of different complexities - the diminishing returns effect is more prominent for more complex datasets but still observable for simpler ones. It is interesting to note that, in absolute terms, more complex problems seem to benefit more from larger models. Further experiments are required to determine if this is due to the limited effective capacity \cite{rethinking} of deep networks, and how it changes with network scaling. In any case, the log-linear trends in Fig. \ref{fig:Results} suggest that it may not be feasible to scale ResNet to solve very large classification problems with potentially hundreds of thousands of categories.

Our empirical approach has a few shortcomings. First, the models have been trained for only 90 epochs using a less robust hyperparameter search. However, as we have kept training process the same among all experiments, we believe the observed trends in Fig. \ref{fig:Results} would persist for more finely tuned models. 

Another shortcoming of our study is that the maximum number of classes (3,761) is insufficient to observe behavior in Region III in Fig \ref{fig:hypothesis}. However, when $N$ is kept constant, as in Fig.~\ref{fig:Results}(b,d), we observe different architectures starting to converge in a low-performance regime, giving some indication that our hypothesis may be correct.

\end{document}